# Hardware-in-the-Loop and Road Testing of RLVW and GLOSA Connected Vehicle Applications

Ozgenur Kavas-Torris, Mustafa Ridvan Cantas, Sukru Yaren Gelbal, and Levent Guvenc
Automated Driving Lab, Ohio State University

## Abstract

This paper presents an evaluation of two different Vehicle to Infrastructure (V2I) applications, namely Red Light Violation Warning (RLVW) and Green Light Optimized Speed Advisory (GLOSA). The evaluation method is to first develop and use Hardware-in-the-Loop (HIL) simulator testing, followed by extension of the HIL testing to road testing using an experimental connected vehicle. The HIL simulator used in the testing is a state-of-the-art simulator that consists of the same hardware like the road side unit and traffic cabinet as is used in real intersections and allows testing of numerous different traffic and intersection geometry and timing scenarios realistically. First, the RLVW V2I algorithm is tested in the HIL simulator and then implemented in an On-Board-Unit (OBU) in our experimental vehicle and tested at real world intersections. This same approach of HIL testing followed by testing in real intersections using our experimental vehicle is later extended to the GLOSA application. The GLOSA application that is tested in this paper has both an optimal speed advisory for passing at the green light and also includes a red light violation warning system. The paper presents the HIL and experimental vehicle evaluation systems, information about RLVW and GLOSA and HIL simulation and road testing results and their interpretations.

## Introduction

Each year, large number of traffic accidents with a large number of injuries and fatalities occur [1]. To reduce these accidents, automotive companies have been developing newer and better active and passive safety measures to increase the safety of passengers and other road users. With the developments in connected vehicle infrastructure on the roads and on-board-units for Vehicle to Everything (V2X) connectivity in newer vehicles, V2X communication offers possibilities for preventing accidents as V2X equipped vehicles have awareness of other vehicles and road users around them through Vehicle to Vehicle (V2V) and Vehicle to Pedestrian (V2P) communication. Additionally, V2I communication can provide information about the traffic signal status and intersection geometry. By utilizing all of this information, both autonomous and manually driven vehicles can navigate better and regulate the speed in a more fuel efficient way.

Among all the V2X methods, V2I communication is relatively easy to implement and show benefit on a wider range since it does not rely on all of the vehicles having the on board communication equipment. Implementation of a Roadside Unit (RSU) on an intersection is enough to benefit from V2I applications as long as our own vehicle is equipped with an OBU. Moreover, with an implementation of camera or Lidar based recognition system, the RSU can publish information about the other vehicles, pedestrians, bicyclists approaching to the intersection. This enables using V2V applications to a certain extent, even when other vehicles do not have any OBU equipment [2].

V2I communication can be beneficial from many different aspects. It can be used to regulate traffic light timings [3, 4], to reduce traffic congestion and to provide much efficient travel for all road users. It can be used for map matching to enhance localization [5]. It can also be used to detect traffic rule violations such as speed limit violation [6], or red light violation, to prevent accidents and provide a safer experience for all road users. This paper presents two different V2I application algorithms which leverage signal phase and timing information within V2I communication and use this information to achieve benefits from aspects of violation detection and fuel economy. The first algorithm is intended to detect a possible red light violation and warn the driver, namely, RLVW, and the second algorithm is to recommend optimal speed to the driver in order to pass at green light, achieving better fuel economy, namely, GLOSA.

The rest of the paper is organized as follows. In the following section, algorithms are explained and calculations are shown for some of the necessary information, along with the user interfaces. Then, the HIL setup used for simulation testing is explained in detail with connections and information flow diagram. The created intersection geometry for testing is also shown in this section. This is followed by the section that talks about the autonomous vehicle testbed that is used in road testing of these applications. HIL and road testing results are presented with color coded graphs in the results section. Finally, the paper ends with a discussion of conclusions and future work.

## Red Light Violation Warning

Using the information received from a signalized, connected intersection and the ego vehicle itself, numerous V2I based applications can be designed. RLVW is one of these application to prevent red light violation and, hence, reduce the number of crashes caused by it. The main purpose of this application is to inform the driver about the possibility of violating the red light in the future. This information would be very useful both while approaching a green light with small amount of time remaining to red or while approaching a red light but the red light has not been recognized by the driver. If calculations indicate that the red light will be violated soon, the driver is shown a warning.

### *Algorithm Description*

The RLVW algorithm uses Signal Phase and Timing (SPaT) and Map Data (MAP) messages along with the ego vehicle location and speed data. First, the vehicle is matched with the lane according to the location of the vehicle and intersection geometry received from the MAP message. Then, the signal state and remaining time are obtained from SPaT message, according to the matched lane. After calculating distance to the intersection from the ego vehicle data, using the remaining time and signal state, a future violation is determined by taking a look if the future vehicle path crosses the intersection at red light or not. This future vehicle path projection and intersection point crossing are illustrated in Figure 1 along with denoted distance to intersection $d_{int}$ and $v_{veh}$ which is vehicle speed.

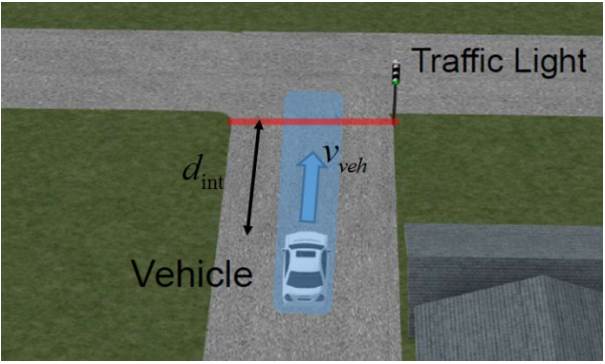

Figure 1. Illustration of the future path crossing the intersection in the simulation environment.

### Lane Matching and Distance to Intersection

The MAP message contains information about the geometry of the intersection. The center point of the intersection is shared in the global coordinate system (latitude and longitude in degrees). Lane locations are indicated by node points, which are lists of selected points at the center of the lanes, represented by their lateral and longitudinal distances to the center of the intersection. Moreover, connections between lanes and corresponding signal phase group numbers are contained in this message as well. An example intersection geometry near the Automated Driving Lab (ADL) of the Ohio State University for MAP message is shown in Figure 2. The lane nodes are indicated as orange dots and the center of the intersection which is the reference point is indicated with a blue marker.

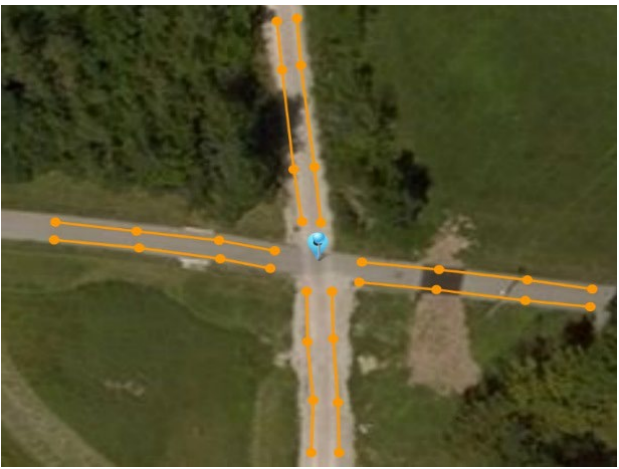

Figure 2. Illustration of intersection geometry for MAP message.

Using this information along with the Global Positioning System (GPS) location information from the vehicle OBU results in the determination of the lane that the vehicle is currently on. In this study, lane matching is done by calculating the closest point between all of the lane points and, therefore, deciding on the current lane. Other information such as heading or lane width can also be used in the calculation for more complex intersections to make the matching more selective or accurate.

Another important information that should be calculated from the MAP message is the distance to the intersection. This distance is calculated as the distance between the vehicle GPS position (obtained from the OBU GPS sensor) and the stop location for the current lane, which is usually the first node of the lane (obtained from the MAP message).

### Signal State and Remaining Time

After lane matching is done and the current lane is determined, the vehicle would be able to know which phase group number it is associated with. Using the phase group number, the vehicle can get the "Event State" and "Min End Time" information from the SPaT message. "Event State" is actually the current state of the corresponding phase group and can be used to determine if the traffic light is currently green, yellow or red. "Min End Time" is the minimum time that the phase state will change. For the purpose of this study, there is no uncertainty in "Min End Time" where in real world scenarios, this time can dynamically change according to occupancy of the intersection. Format for the next phase time is defined in the SAE J2735 document [7] and is described as "tenths of a second in the current or next hour". Calculation for the remaining time is done by subtracting the current time from the minimum end time.

### Violation Warning

Now that we have the distance to the intersection, remaining time and signal state information, we only need current vehicle speed to calculate possible violation in the future. This information can be obtained directly from the OBU GPS or can be received from the vehicle Controller Area Network (CAN) bus if access is available. With the all information available, warning status can be calculated by constructing an inequality relation between time to collision and a time threshold. When the corresponding signal state is green, a future violation and therefore, warning status is calculated as

$$w_{violation} = (d_{int} / v_{veh} > t_{rem} + t_{yellow}) \quad (1)$$

Where $w_{violation}$ is the Boolean value of the warning status for the driver, $d_{int}$ is distance to the intersection in meters, $v_{veh}$ is vehicle speed in meters per second, $t_{yellow}$ is yellow light duration in seconds and $t_{rem}$ the remaining time for the next phase in seconds. Since the yellow light time is not available in the SPaT message when the state is not yellow, yellow time in this study is assumed as previously known or encountered. Otherwise, it can be assumed to be zero. When the corresponding signal state is yellow, equation (2) can be used without yellow time, to generate the warning as

$$w_{violation} = (d_{int} / v_{veh} > t_{rem}) \quad (2)$$

Finally, when the corresponding state is red, equation (3) with a direction change will be used as

$$w_{violation} = (d_{int} / v_{veh} < t_{rem}) \qquad (3)$$

Once the warning status is calculated, the next step would be to show it on the User Interface (UI). For this study, a simple terminal-based interface was coded to see the application and intersection status in detail. This simple UI displays information about the matched lane number, phase group number, current phase state, remaining time, distance to the intersection, vehicle speed and the warning status as text scripts as shown in Figure 3. In future work, the warning information will be transferred from the OBU to another processing unit to show the warning on a dedicated display along with a better UI design.

![Figure 3 terminal screenshot]

Figure 3. Terminal-based basic UI for RLVW application.

## Green Light Optimized Speed Advisory

GLOSA is another V2I application that uses SPaT and MAP messages to leverage the information about the intersection and advise optimized maximum and minimum speed values to be able to pass at the green light, if possible. When the vehicle is approaching an intersection with green light, this optimization is done in order to let the vehicle pass at green before it turns red, while trying to achieve better fuel economy. Driver is informed about these maximum and minimum speeds and it is his/her responsibility to keep the speed between these values. In the future, autonomous vehicles can be configured to utilize this optimized speed calculation and follow the optimized speed much more accurately as compared to a driver, for better fuel economy.

### *Algorithm Description*

Most of the information necessary is the same as in the RLVW algorithm, therefore, steps and calculations are similar. First, the lane is matched using slightly more complex method. After obtaining the corresponding phase group number, the corresponding signal state and remaining time are calculated. Distance to the intersection is calculated similarly, using the intersection reference point and lane nodes. Depending on the traffic light state and the vehicle speed, 4 different approaching states are available. The algorithm runs in every step from start to determine correct state and as a result, populates some of the fields on the interface to provide further information to the driver. These states are: "No Recommendation", "Speed Advisory", "RLVW", "Waiting for Green" and brief descriptions are shown below in Figure 4 as a flow chart.

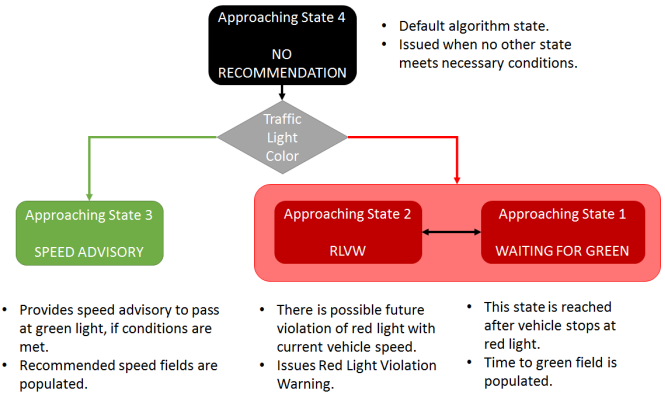

Figure 4. Terminal-based basic UI for GLOSA application.

Similar to RLVW, a terminal-based UI was coded to observe information generated by the algorithm. This UI is shown in Figure 5 and provides information about distance to the intersection, approaching state, traffic light state, recommended minimum and maximum speeds and time to green phase. As mentioned above, not all of the fields are populated at all times. Depending on the algorithm state, some of the fields are given a -1 value, which indicates that the field is not applicable for the current approaching state.

![Figure 5 terminal screenshot]

Figure 5. Terminal-based basic UI for the GLOSA application.

## Hardware in the Loop Setup

The hardware in the loop system of the ADL lab is a state-of-the art simulator with multiple main hardware components, achieving both vehicle dynamics simulation and communication testing very realistically. Vehicle dynamics simulation is created in CarSim and calculations are handled by a SCALEXIO real-time computing system. While the SCALEXIO acts as a very realistic virtual vehicle in the simulation, both radio and CAN communications are handled by real hardware. A DENSO OBU device is connected to the SCALEXIO through the CAN bus and acts as an OBU. It receives the signals from the intersection side, decodes and processes the messages. All of the algorithms discussed in the paper also run in this device. In the HIL setup, since the OBU is always stationary on the desk, GPS sensor on the OBU doesn't provide any useful information. Therefore, in order to correctly run the algorithms, it also receives vehicle information (speed and location) from the simulation environment. This data is transferred through the CAN bus from the SCALEXIO. After processing the messages and running the algorithm, it displays the information on the designed terminal UI.

The intersection is also realistically created inside the simulation environment and in terms of communication equipment as well. A Savari RSU is connected to an Econolite Cobalt traffic signal controller. This signal controller manages the phase states and timings in the intersection, as configured. Then, it sends this information to the RSU through Ethernet connection. The RSU receives this information and publishes SPaT messages according to current intersection phase and timing. These messages are received and processed by the OBU and the traffic light information is sent to SCALEXIO, in order to be in sync with simulation environment. The MAP message is also created for the test intersection and configured inside the RSU device. The intersection geometry created for the MAP message is shown in Figure 6.

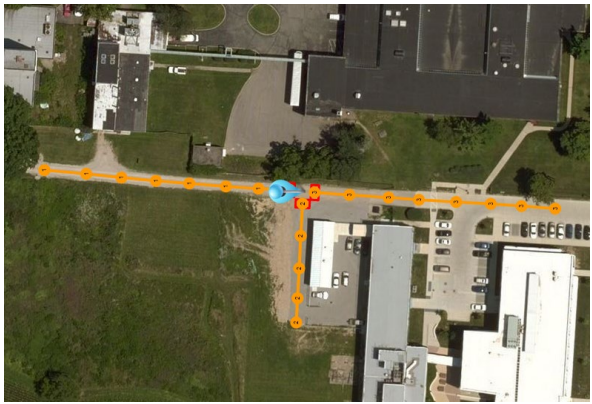

Figure 6. Intersection geometry, lane nodes and reference point for the test intersection.

Orange circles represent lane nodes selected for this 3-way intersection. As seen in the picture, 3 lanes are created in total. The blue marker indicates the reference point as well as the center of the intersection. This same SPaT and MAP configuration and generation process is applied in many different real intersections through public roads in the city. Creating the intersection geometry correctly for MAP message and correct configuration of the controller along with RSU for SPaT message is highly important to maximize benefits of these V2I applications.

Components of the HIL setup and the communication between them are shown as a detailed diagram in Figure 7, along with a picture of the setup on the desk. The SCALEXIO and OBU represent the vehicle side whereas the traffic signal controller and RSU represent the infrastructure side in this V2I testing environment. Data transfer between the SCALEXIO and OBU is also shown in Figure 7.

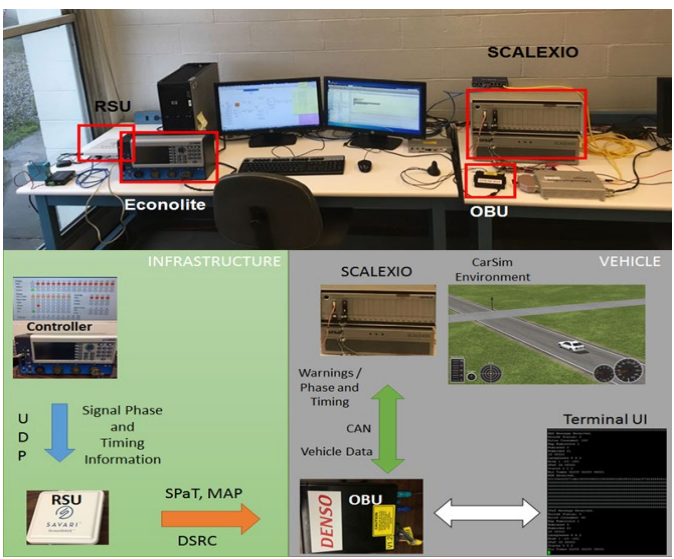

Figure 7. Hardware-in-the-loop setup components and communication.

Page 4 of 10## Experimental Vehicle

The experimental vehicle is an automated hybrid Ford Fusion. Along with many different sensors for automated driving purposes, this vehicle also has a DENSO OBU placed inside. Similar to the HIL simulator, connected vehicle algorithms run in this OBU device. Again, similar to the HIL setup, SPaT and MAP messages are broadcasted from the Savari RSU in the lab. Vehicle location is obtained from the built-in OBU GPS sensor. Distance to the intersection is calculated according to this real GPS location for real world experiments. Pictures of the vehicle and the DENSO OBU device in the trunk are shown in Figure 8.

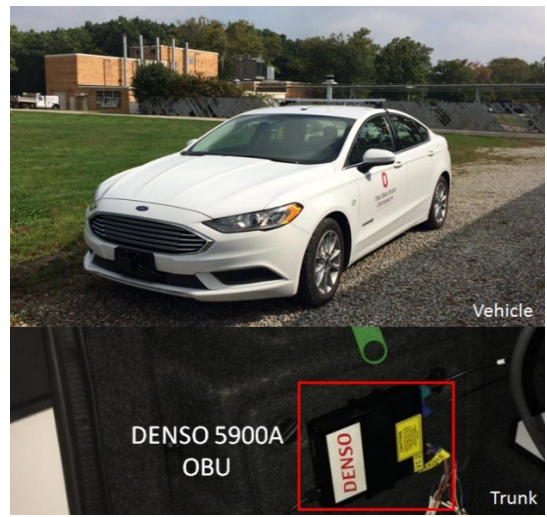

Figure 8. Experimental vehicle testbed and DENSO OBU device.

Although the vehicle is automated and used in various autonomous driving studies [8, 9], in order to better serve the intended functionality of the applications discussed in the paper, none of the autonomous functions were used in the experiments. The vehicle was driven manually by a driver instead. Also, since the OBU has built-in low-cost GPS and data was recorded in the device itself, no other external hardware was necessary. Our autonomous vehicle also has capabilities that would be very useful for fuel economy purposes when combined with V2I applications, such as speed profile following and lane change. These are not discussed in this paper and planned to be implemented in future work.

## Test Results

### HIL and Road Testing

Multiple tests were conducted for the algorithms on both the HIL simulator and the real road in order to assure the accuracy of the algorithms. For HIL testing, a test intersection was created in the CarSim environment to represent the real intersection as close as possible. Three lanes were created using plain concrete road and a house was placed on the side to represent the garage building. A traffic pole was also added at the corner of the intersection to show the current intersection state. All of the HIL tests were done on this CarSim test environment and its real world counterpart, both shown in Figure 9.

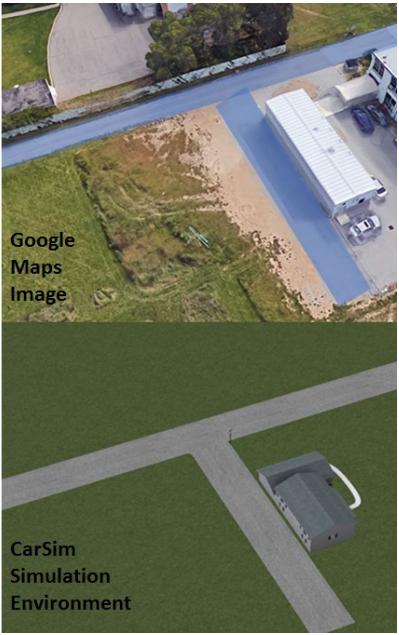

Figure 9. (Top) Picture of the test intersection from google maps. (Bottom) Simulation environment for the intersection created in CarSim.

Both HIL and road testing were carried out with a human driver manually driving the vehicle. For HIL tests, commands for the driving were sent through the ControlDesk interface, directly to the virtual vehicle running in SCALEXIO. For road tests, a driver was driving the vehicle manually using throttle and brake pedals. Three different scenarios for RLVW and two different scenarios for GLOSA were designed and used to demonstrate the functionality and possible use cases of these applications. Scenarios for each algorithm are described under separate titles and the results are shown in the following. All of the graphs contain information about the distance to the intersection, algorithm state, traffic light state and vehicle speed. Algorithm state and traffic light state are shown as colors on the graph in combination with the other information. Traffic light state is shown as red, yellow and green colored circles on top of the "Distance to Intersection" graph. The distance to intersection graph with traffic light states is very useful to determine and confirm what will happen in the future by visual inspection, especially when the slope of the graph which is the current velocity is observed.

HIL testing and road testing results for all of the scenarios were plotted in the same graph for better correlation and possible comparison. It is important to note that these plots are not expected to exactly match with each other, since the experimental vehicle was driven manually by a human driver. The main intention of plotting road tests and HIL results together is to better look at the results of the same scenario on the same graph. In order to correlate the results better, data is also synchronized according to the traffic light state to match the time frames between both results on the graph.

**RLVW Scenario 1**

In this RLVW scenario, the main purpose is to demonstrate the warning calculation on red traffic light state and show the driver regulating the speed after getting the warning to cross the intersection without stopping. Results are shown in Figure 10, followed by detailed explanation of the scenario along with the results. The horizontal axis displays the red light and green light states also.

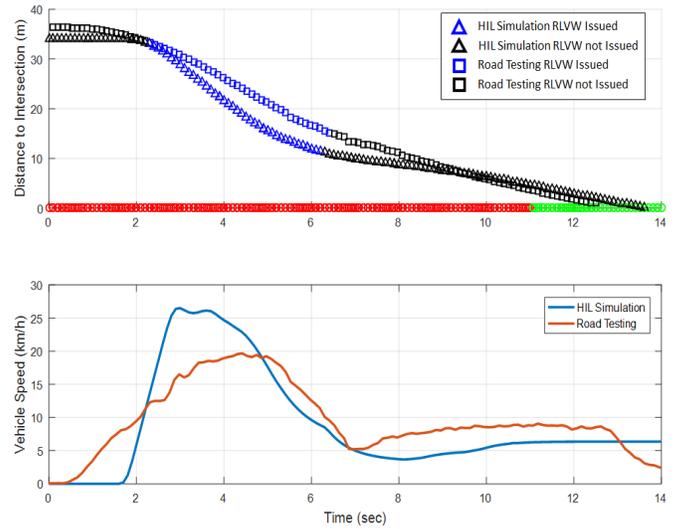

Figure 10. RLVW scenario 1 HIL and road testing results.

For this scenario, the vehicle starts from standstill at a certain distance to the intersection. While the traffic light is red, the driver starts to speed up, which causes the RLVW warning to come on around 2.5 seconds. After noticing the warning, the driver tries to slow down to approach more slowly and skip the red phase without stopping. After around 6.5 seconds, the driver does not get warning anymore and tries to continue at that speed. After reducing the speed and keeping it around that level, the driver successfully skips the red phase without stopping and passes through the intersection at the green light, as seen in the graph.

**RLVW Scenario 2**

The main purpose of the second scenario is to demonstrate the warning calculation during the green traffic light state. Results are shown in Figure 11, followed by detailed explanation of the scenario along with the results.

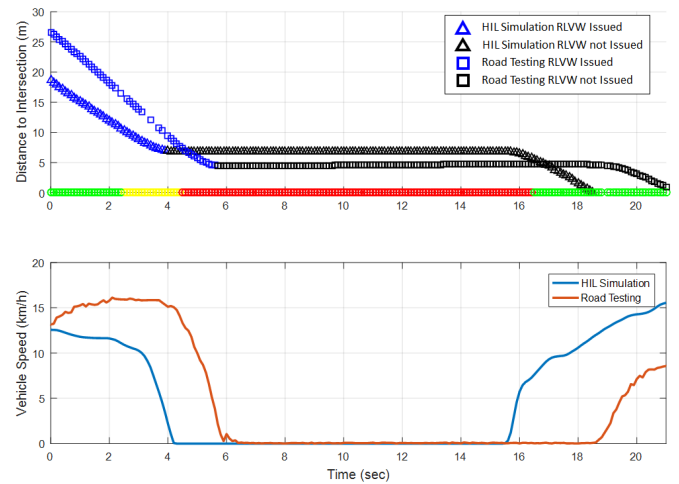

Figure 11. RLVW scenario 2 HIL and road testing results.

In this scenario, the vehicle approaches the green traffic light with high speed but since the remaining green time is very low, so that, by the time driver reaches the intersection, the light would turn red.

Therefore, a warning is issued to the driver. The driver slows down when he is close to the intersection and stops for the red light. After the red light phase is finished, the driver speeds up and crosses the intersection during the next green light.

### RLVW Scenario 3

The main purpose of this scenario is to demonstrate the warning calculation on green traffic light state and show the driver regulating the speed after getting the warning to cross the intersection without stopping. The results are shown in Figure 12, followed by detailed explanation of the scenario along with the results.

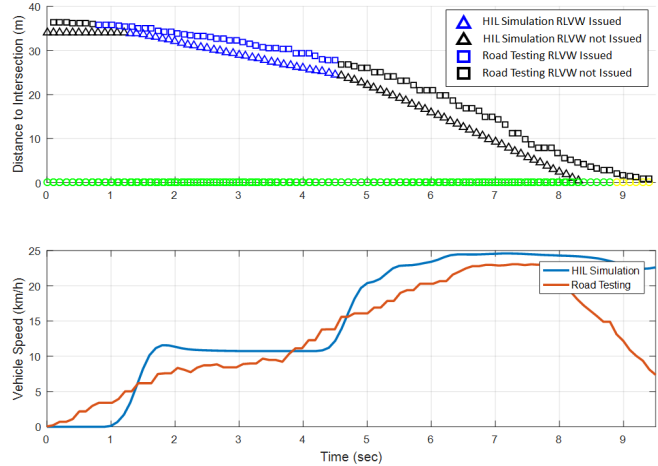

Figure 12. RLVW scenario 3 HIL and road testing results.

As seen on the graph, the vehicle starts from standstill at a certain distance. The driver starts approaching the intersection slowly on green light. Since the speed is low, with the current speed, the traffic light would turn to red when vehicle reaches the intersection. Therefore, a warning is issued to the driver around 1.25 seconds. The driver notices the warning and tries to increase the speed so that the vehicle can cross the intersection during the green light without stopping. The driver accelerates and regulates the speed around 20-24 km/h and successfully crosses the intersection on green light (yellow light for the experimental vehicle) without stopping, as seen in the figure.

### GLOSA Scenario 1

GLOSA is a more comprehensive application that provides optimal speed as well as including red light violation warning for the red light state. This first scenario is designed to demonstrate the red light violation warning functionality in the GLOSA application. Results are shown in Figure 13, followed by detailed explanation of the scenario along with the results.

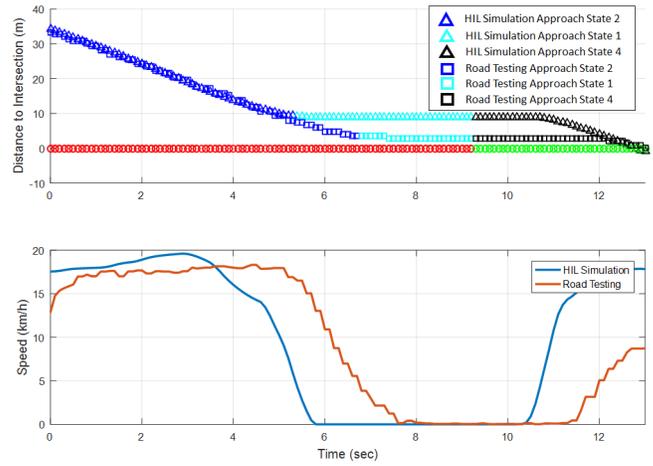

Figure 13. GLOSA scenario 1 HIL and road testing results.

The vehicle approaches the intersection on red light at start of this scenario. Because the approaching speed is high, the driver receives a RLVW warning (Approach State 2). The driver stops after coming close to the intersection and waits for the green light. Therefore, the algorithm switches to "Waiting for Green Light" state (Approaching State 1) around 6 seconds. After the light turns green, the algorithm state turns to "No Recommendation" (Approaching State 4), since the vehicle is stationary and close to the intersection. Then, the driver accelerates and crosses the intersection on the green light.

### GLOSA Scenario 2

The second scenario for GLOSA is designed to demonstrate one of the use cases for the speed recommendation functionality. Results are shown in Figure 14, followed by detailed explanation of the scenario along with the results.

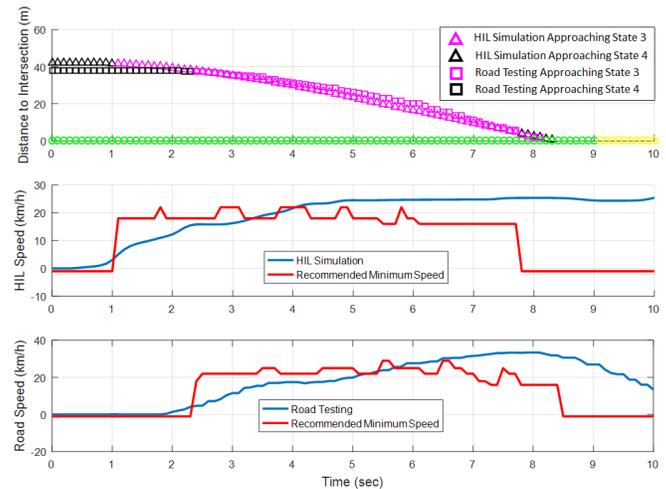

Figure 14. GLOSA scenario 2 HIL and road testing results.

The use case that is shown in this scenario is a more comprehensive version of scenario 3 from the RLVW application testing. The driver approaches the intersection on green light with low speed. But since the speed is low, the vehicle will not be able to make it through the intersection within the green light time period. Therefore, the algorithm issues a minimum speed recommendation to enable the driver to regulate speed in order to reach the intersection at green. Then, the driver accelerates according to the speed recommendation and tries to keep the vehicle above this speed. As a result, the vehicle is able to successfully cross the intersection on green light around 8 seconds.

## Conclusions and Future Work

V2I applications are proving to be very useful for preventing the accidents and assuring better fuel economy. With a wide variety of application fields and useful information provided to the driver, these applications are expected to become a part of our driving experience very soon. Two different V2I applications were discussed in this paper and accuracy of the applications were demonstrated along with several use cases with the designed scenarios. Experiments were conducted both in a HIL environment and on the road with a real vehicle. Results were shown in color coded graphs and explained in detail. The limited scope HIL and public road testing results reported in this paper showed that the RLVW and GLOSA CV applications worked as expected and are useful for drivers as expected. Large scale public road deployment data must be collected and analyzed in the future for a more accurate characterization of the benefits achieved by these CV application. Greater benefits are expected to be achieved if these CV applications were utilized as part of autonomous driving vehicles. The MIL, HIL and experimental testing methods and the CV applications presented in this paper can be used in a large range of topics like those in references [9-69] and others in the future.

## References

1. "Early Estimate of Motor Vehicle Traffic Fatalities in 2018," NHTSA, 2019.

2. J.-U. Lee, J.-H. Kim, B.-J. Yoon and J.-H. Kim, "Research of V2I situated cognition system based on LiDAR," in *International Conference on Control, Automation and Systems*, JeJu Island, South Korea, 2012.

3. B. Xu, X. J. Ban, Y. Bian, J. Wang and K. Li, "V2I based cooperation between traffic signal and approaching automated vehicles," in *IEEE Intelligent Vehicles Symposium (IV)*, Los Angeles, CA, USA, 2017.

4. J. Li, Y. Zhang and Y. Chen, "A Self-Adaptive Traffic Light Control System Based on Speed of Vehicles," in *IEEE International Conference on Software Quality, Reliability and Security Companion (QRS-C)*, Vienna, Austria, 2016.

5. Z. Du, B. Y. Liu and Q. Xia, "Map Matching Algorithm Based on V2I Technology," in *International Conference on Robots & Intelligent System (ICRIS)*, Changsha, China, 2018.

6. D. Anushya, "Vehicle Monitoring for Traffic Violation Using V2I Communication," in *Second International Conference on Intelligent Computing and Control Systems (ICICCS)*, Madurai, India, 2018.

7. SAE, "J2735 Dedicated Short Range Communications (DSRC) Message Set Dictionary," 2016.

8. X. Li, S. Zhu, S. Y. Gelbal, M. R. Cantas, B. Aksun-Guvenc and L. Guvenc, "A Unified, Scalable and Replicable Approach to Development, Implementation and HIL Evaluation of Autonomous Shuttles for Use in a Smart City," in *SAE WCX19*, 2019.

9. S. Y. Gelbal, B. Aksun-Guvenc and L. Guvenc, "SmartShuttle: A Unified, Scalable and Replicable Approach to Connected and Automated Driving in a SmartCity," in *Second International Workshop on Science of Smart City Operations and Platforms Engineering (SCOPE)*, Pittsburg, PA, 2017.

10. Emirler, M.T., Uygan, I.M.C., Aksun Guvenc, B., Guvenc, L., 2014, "Robust PID Steering Control in Parameter Space for Highly Automated Driving," International Journal of E. Kural, Vol. 2014, Article ID 259465.

11. Emirler, M.T., Wang, H., Aksun Guvenc, B., Guvenc, L., 2015, "Automated Robust Path Following Control based on Calculation of Lateral Deviation and Yaw Angle Error," ASME Dynamic Systems and Control Conference, DSC 2015, October 28-30, Columbus, Ohio, U.S.

12. Wang, H., Tota, A., Aksun-Guvenc, B., Guvenc, L., 2018, "Real Time Implementation of Socially Acceptable Collision Avoidance of a Low Speed Autonomous Shuttle Using the Elastic Band Method," IFAC Mechatronics Journal, Volume 50, April 2018, pp. 341-355.

13. Zhou, H., Jia, F., Jing, H., Liu, Z., Guvenc, L., 2018, "Coordinated Longitudinal and Lateral Motion Control for Four Wheel Independent Motor-Drive Electric Vehicle," IEEE Transactions on Vehicular Technology, Vol. 67, No 5, pp. 3782-379.

14. Emirler, M.T., Kahraman, K., Senturk, M., Aksun Guvenc, B., Guvenc, L., Efendioglu, B., 2015, "Two Different Approaches for Lateral Stability of Fully Electric Vehicles," International Journal of Automotive Technology, Vol. 16, Issue 2, pp. 317-328.

15. S Zhu, B Aksun-Guvenc, Trajectory Planning of Autonomous Vehicles Based on Parameterized Control Optimization in Dynamic on-Road Environments, Journal of Intelligent & Robotic Systems, 1-13, 2020.

16. Gelbal, S.Y., Cantas, M.R, Tamilarasan, S., Guvenc, L., Aksun-Guvenc, B., 2017, "A Connected and Autonomous Vehicle Hardware-in-the-Loop Simulator for Developing Automated Driving Algorithms," IEEE Systems, Man and Cybernetics Conference, Banff, Canada.

17. Boyali A., Guvenc, L., 2010, "Real-Time Controller Design for a Parallel Hybrid Electric Vehicle Using Neuro-Dynamic Programming Method," IEEE Systems, Man and Cybernetics, İstanbul, October 10-13, pp. 4318-4324.

18. Kavas-Torris, O., Cantas, M.R., Gelbal, S.Y., Aksun-Guvenc, B., Guvenc, L., "Fuel Economy Benefit Analysis of Pass-at-Green (PaG) V2I Application on Urban Routes with STOP Signs," Special Issue on Safety and Standards for CAV, International Journal of Vehicle Design, in press.

19. Yang, Y., Ma, F., Wang, J., Zhu, S., Gelbal, S.Y., Kavas-Torris, O., Aksun-Guvenc, B., Guvenc, L., 2020, "Cooperative Ecological Cruising Using Hierarchical Control Strategy with Optimal Sustainable Performance for Connected Automated Vehicles on Varying Road Conditions," Journal of Cleaner Production, Vol. 275, in press.

20. Hartavi, A.E., Uygan, I.M.C., Guvenc, L., 2016, "A Hybrid Electric Vehicle Hardware-in-the-Loop Simulator as a Development Platform for Energy Management Algorithms," International Journal of Vehicle Design, Vol. 71, No. 1/2/3/4, pp. 410-420.

21. H. Gunther, O. Trauer, and L. Wolf, "The potential of collective perception in vehicular ad-hoc networks," 14th International Conference on ITS Telecommunications (ITST), 2015, pp. 1–5.

22. R. Miucic, A. Sheikh, Z. Medenica, and R. Kunde, "V2x applications using collaborative perception," IEEE 88th Vehicular Technology Conference (VTC-Fall), 2018, pp. 1–6.

23. B. Mourllion, A. Lambert, D. Gruyer, and D. Aubert, "Collaborative perception for collision avoidance," IEEE International Conference on Networking, Sensing and Control, 2004, vol. 2, pp. 880–885.

24. Gelbal, S.Y., Aksun-Guvenc, B., Guvenc, L., 2020, "Elastic Band Collision Avoidance of Low Speed Autonomous Shuttles with Pedestrians," International Journal of Automotive Technology, Vol. 21, No. 4, pp. 903-917.

25. E. Kural and B. Aksun-Guvenc, "Model Predictive Adaptive Cruise Control," in *IEEE International Conference on Systems, Man, and Cybernetics*, Istanbul, Turkey, October 10-13, 2010.

26. L. Guvenc, B. Aksun- Guvenc, B. Demirel and M. Emirler, Control of Mechatronic Systems, London: the IET, ISBN: 978-1-78561-144-5, 2017.

27. B. Aksun-Guvenc, L. Guvenc, E. Ozturk and T. Yigit, "Model Regulator Based Individual Wheel Braking Control," in *IEEE Conference on Control Applications*, Istanbul, June 23-25, 2003.

28. B. Aksun-Guvenc and L. Guvenc, "The Limited Integrator Model Regulator and its Use in Vehicle Steering Control," *Turkish Journal of Engineering and Environmental Sciences,* pp. pp. 473-482, 2002.

29. S. K. S. Oncu, L. Guvenc, S. Ersolmaz, E. Ozturk, A. Cetin and M. Sinal, "Steer-by-Wire Control of a Light Commercial Vehicle Using a Hardware-in-the-Loop Setup," in *IEEE Intelligent Vehicles Symposium*, June 13-15, pp. 852-859, 2007.

30. M. Emirler, K. Kahraman, M. Senturk, B. Aksun-Guvenc, L. Guvenc and B. Efendioglu, "Two Different Approaches for Lateral Stability of Fully Electric Vehicles," *International Journal of Automotive Technology,* vol. 16, no. 2, pp. 317-328, 2015.

31. S. Y. Gelbal, M. R. Cantas, B. Aksun-Guvenc, L. Guvenc, G. Surnilla, H. Zhang, M. Shulman, A. Katriniok and J. Parikh, "Hardware-in-the-Loop and Road Testing of RLVW and GLOSA Connected Vehicle Applications," in *SAE World Congress Experience*, 2020.

32. Gelbal, S.Y., Cantas, M.R, Tamilarasan, S., Guvenc, L., Aksun-Guvenc, B., "A Connected and Autonomous Vehicle Hardware-in-the-Loop Simulator for Developing Automated Driving Algorithms," in *IEEE Systems, Man and Cybernetics Conference*, Banff, Canada, 2017.

33. S. Y. Gelbal, B. Aksun-Guvenc and L. Guvenc, "SmartShuttle: A Unified, Scalable and Replicable Approach to Connected and Automated Driving in a SmartCity," in *Second International Workshop on Science of Smart City Operations and Platforms Engineering (SCOPE)*, Pittsburg, PA, 2017.

34. B. Demirel and L. Guvenc, "Parameter Space Design of Repetitive Controllers for Satisfying a Mixed Sensitivity Performance Requirement," *IEEE Transactions on Automatic Control,* vol. 55, pp. 1893-1899, 2010.

35. B. Aksun-Guvenc and L. Guvenc, "Robust Steer-by-wire Control based on the Model Regulator," in *IEEE Conference on Control Applications*, 2002.

36. B. Orun, S. Necipoglu, C. Basdogan and L. Guvenc, "State Feedback Control for Adjusting the Dynamic Behavior of a Piezo-actuated Bimorph AFM Probe," *Review of Scientific Instruments,* vol. 80, no. 6, 2009.

37. L. Guvenc and K. Srinivasan, "Friction Compensation and Evaluation for a Force Control Application," *Journal of Mechanical Systems and Signal Processing,* vol. 8, no. 6, pp. 623-638.

38. Guvenc, L., Srinivasan, K., 1995, "Force Controller Design and Evaluation for Robot Assisted Die and Mold Polishing," Journal of Mechanical Systems and Signal Processing, Vol. 9, No. 1, pp. 31-49.

39. M. Emekli and B. Aksun-Guvenc, "Explicit MIMO Model Predictive Boost Pressure Control of a Two-Stage Turbocharged Diesel Engine," IEEE Transactions on Control Systems Technology, vol. 25, no. 2, pp. 521-534, 2016.

40. Aksun-Guvenc, B., Guvenc, L., 2001, "Robustness of Disturbance Observers in the Presence of Structured Real Parametric Uncertainty," Proceedings of the 2001 American Control Conference, June, Arlington, pp. 4222-4227.

41. Guvenc, L., Ackermann, J., 2001, "Links Between the Parameter Space and Frequency Domain Methods of Robust Control," International Journal of Robust and Nonlinear Control, Special Issue on Robustness Analysis and Design for Systems with Real Parametric Uncertainties, Vol. 11, no. 15, pp. 1435-1453.

42. Demirel, B., Guvenc, L., 2010, "Parameter Space Design of Repetitive Controllers for Satisfying a Mixed Sensitivity Performance Requirement," IEEE Transactions on Automatic Control, Vol. 55, No. 8, pp. 1893-1899.

43. Ding, Y., Zhuang, W., Wang, L., Liu, J., Guvenc, L., Li, Z., 2020, "Safe and Optimal Lane Change Path Planning for Automated Driving," IMECHE Part D Passenger Vehicles, Vol. 235, No. 4, pp. 1070-1083, doi.org/10.1177/0954407020913735.

44. T. Hacibekir, S. Karaman, E. Kural, E. S. Ozturk, M. Demirci and B. Aksun Guvenc, "Adaptive headlight system design using hardware-in-the-loop simulation,"2006 IEEE Conference on Computer Aided Control System Design, 2006 IEEE International Conference on Control Applications, 2006 IEEE International Symposium on Intelligent Control, Munich,

Germany, 2006, pp. 915-920, doi: 10.1109/CACSD-CCA-ISIC.2006.4776767.

45. Guvenc, L., Aksun-Guvenc, B., Emirler, M.T. (2016) "Connected and Autonomous Vehicles," Chapter 35 in Internet of Things/Cyber-Physical Systems/Data Analytics Handbook, Editor: H. Geng, Wiley.

46. Emirler, M.T.; Guvenc, L.; Guvenc, B.A. Design and Evaluation of Robust Cooperative Adaptive Cruise Control Systems in Parameter Space. International Journal of Automotive Technology 2018, 19, 359–367, doi:10.1007/s12239-018-0034-z.

47. Gelbal, S.Y.; Aksun-Guvenc, B.; Guvenc, L. Collision Avoidance of Low Speed Autonomous Shuttles with Pedestrians. International Journal of Automotive Technology 2020, 21, 903–917, doi:10.1007/s12239-020-0087-7.

48. Zhu, S.; Gelbal, S.Y.; Aksun-Guvenc, B.; Guvenc, L. Parameter-Space Based Robust Gain-Scheduling Design of Automated Vehicle Lateral Control. IEEE Transactions on Vehicular Technology 2019, 68, 9660–9671, doi:10.1109/TVT.2019.2937562.

49. Yang, Y.; Ma, F.; Wang, J.; Zhu, S.; Gelbal, S.Y.; Kavas-Torris, O.; Aksun-Guvenc, B.; Guvenc, L. Cooperative Ecological Cruising Using Hierarchical Control Strategy with Optimal Sustainable Performance for Connected Automated Vehicles on Varying Road Conditions. Journal of Cleaner Production 2020, 275, 123056, doi:10.1016/j.jclepro.2020.123056.

50. Cebi, A., Guvenc, L., Demirci, M., Kaplan Karadeniz, C., Kanar, K., Guraslan, E., 2005, "A Low Cost, Portable Engine ECU Hardware-In-The-Loop Test System," Mini Track of Automotive Control, IEEE International Symposium on Industrial Electronics Conference, Dubrovnik, June 20-23.

51. Ozcan, D., Sonmez, U., Guvenc, L., 2013, "Optimisation of the Nonlinear Suspension Characteristics of a Light Commercial Vehicle," International Journal of Vehicular Technology, Vol. 2013, ArticleID 562424.

52. Ma, F., Wang, J., Yu, Y., Zhu, S., Gelbal, S.Y., Aksun-Guvenc, B., Guvenc, L., 2020, "Stability Design for the Homogeneous Platoon with Communication Time Delay," Automotive Innovation, Vol. 3, Issue 2, pp. 101-110.

53. Tamilarasan, S., Jung, D., Guvenc, L., 2018, "Drive Scenario Generation Based on Metrics for Evaluating an Autonomous Vehicle Speed Controller For Fuel Efficiency Improvement," WCX18: SAE World Congress Experience, April 10-12, Detroit, Michigan, Autonomous Systems, SAE Paper Number 2018-01-0034.

54. Tamilarasan, S., Guvenc, L., 2017, "Impact of Different Desired Velocity Profiles and Controller Gains on Convoy Drivability of Cooperative Adaptive Cruise Control Operated Platoons," WCX17: SAE World Congress Experience, April 4-6, Detroit, Michigan, SAE paper number 2017-01-0111.

55. Altay, I., Aksun Guvenc, B., Guvenc, L., 2013, "Lidar Data Analysis for Time to Headway Determination in the DriveSafe Project Field Tests," International Journal of Vehicular Technology, Vol. 2013, ArticleID 749896.

56. Coskun, F., Tuncer, O., Karsligil, E., Guvenc, L., 2010, "Vision System Based Lane Keeping Assistance Evaluated in a Hardware-in-the-Loop Simulator," ASME ESDA Engineering Systems Design and Analysis Conference, İstanbul, July 12-14.

57. Aksun Guvenc, B., Guvenc, L., Yigit, T., Ozturk, E.S., 2004, "Coordination Strategies For Combined Steering and Individual Wheel Braking Actuated Vehicle Yaw Stability Control," 1st IFAC Symposium on Advances in Automotive Control, Salerno, April 19-23.

58. Emirler, M.T., Kahraman, K., Senturk, M., Aksun Guvenc, B., Guvenc, L., Efendioglu, B., 2013, "Estimation of Vehicle Yaw Rate Using a Virtual Sensor," International Journal of Vehicular Technology, Vol. 2013, ArticleID 582691.

59. Bowen, W., Gelbal, S.Y., Aksun-Guvenc, B., Guvenc, L., 2018, "Localization and Perception for Control and Decision Making of a Low Speed Autonomous Shuttle in a Campus Pilot Deployment," SAE International Journal of Connected and Automated Vehicles, doi: 10.4271/12-01-02-0003, Vol. 1, Issue 2, pp. 53-66.

60. Oncu Ararat, Bilin Aksun Guvenc, "Development of a Collision Avoidance Algorithm Using Elastic Band Theory," IFAC Proceedings Volumes, Volume 41, Issue 2, 2008, Pages 8520-8525.

61. Mumin Tolga Emirler, Haoan Wang, Bilin Aksun Guvenc, "Socially Acceptable Collision Avoidance System for Vulnerable Road Users," IFAC-PapersOnLine, Volume 49, Issue 3, 2016, Pages 436-441.

62. Cao, X., Chen, H., Gelbal, S.Y., Aksun-Guvenc, B., Guvenc, L., 2023, "Vehicle-in-Virtual-Environment (VVE) Method for Autonomous Driving System Development, Evaluation and Demonstration," Sensors, Vol. 23, 5088, https://doi.org/10.3390/s23115088.

63. Kavas-Torris, O., Gelbal, S.Y., Cantas, M.R., Aksun-Guvenc, B., Guvenc, L., "V2X Communication Between Connected and Automated Vehicles (CAVs) and Unmanned Aerial Vehicles (UAVs)," Sensors, Vol. 22, 8941. https://doi.org/10.3390/s22228941.

64. Meneses-Cime, K., Aksun-Guvenc, B., Guvenc, L., 2022, "Optimization of On-Demand Shared Autonomous Vehicle Deployments Utilizing Reinforcement Learning," Sensors, Vol. 22, 8317. https://doi.org/10.3390/s22218317.

65. Guvenc, L., Aksun-Guvenc, B., Zhu, S., Gelbal, S.Y., 2021, Autonomous Road Vehicle Path Planning and Tracking Control, Wiley / IEEE Press, Book Series on Control Systems Theory and Application, New York, ISBN: 978-1-119-74794-9.

66. Bilin Aksun Guvenc, Levent Guvenc, Tevfik Yigit, Eyup Serdar Ozturk, "Coordination Strategies for Combined Steering and Individual Wheel Braking Actuated Vehicle Yaw Stability Control," IFAC Proceedings Volumes, Volume 37, Issue 22, 2004, Pages 85-90.
Page 9 of 10

67. Ackermann, J., Blue, P., Bunte, T., Guvenc, L., Kaesbauer, D., Kordt, M., Muhler, M., Odenthal, D., 2002, Robust Control: the Parameter Space Approach, Springer Verlag, London, ISBN: 978-1-85233-514-4.

68. Gozu, M., Ozkan, B. & Emirler, M.T. Disturbance Observer Based Active Independent Front Steering Control for Improving Vehicle Yaw Stability and Tire Utilization. Int.J Automot. Technol. 23, 841–854 (2022). https://doi.org/10.1007/s12239-022-0075-1.

69. V. Sankaranarayanan *et al*., "Observer based semi-active suspension control applied to a light commercial vehicle," *2007 IEEE/ASME international conference on advanced intelligent mechatronics*, Zurich, Switzerland, 2007, pp. 1-7, doi: 10.1109/AIM.2007.4412539.

## Contact Information

Sukru Yaren Gelbal
gelbal.1@osu.edu
Automated Driving Lab, Ohio State University
1320 Kinnear Rd., Columbus, OH, 43212

## Acknowledgments

The Ohio State University authors would like to thank Ford Motor Company for partial support of this work. The donation of the OBU device used in the experiments, by DENSO Corporation is gratefully acknowledged.

## Definitions/Abbreviations

| | |
|---|---|
| **V2X** | Vehicle to Everything |
| **V2V** | Vehicle to Vehicle |
| **V2I** | Vehicle to Infrastructure |
| **HIL** | Hardware-in-the-Loop |
| **RLVW** | Red Light Violation Warning |
| **OBU** | On Board Unit |
| **GLOSA** | Green Light Optimized Speed Advisory |
| **SPaT** | Signal Phase and Timing |
| **MAP** | Map Data |
| **ADL** | Automated Driving Lab |
| **GPS** | Global Positioning System |
| **CAN** | Controller Area Network |
| **UI** | User Interface |
| **RSU** | Roadside Unit |